\documentclass[sigconf]{acmart}
\AtBeginDocument{%
  }

\setcopyright{acmlicensed}
\copyrightyear{2018}
\acmYear{2018}
\acmDOI{XXXXXXX.XXXXXXX}
\acmConference[Conference acronym 'XX]{Make sure to enter the correct
  conference title from your rights confirmation email}{June 03--05,
  2018}{Woodstock, NY}
\acmISBN{978-1-4503-XXXX-X/2018/06}

\usepackage{soul}
\usepackage{enumitem}
\usepackage{flushend}
\usepackage{balance}
\usepackage{algorithm}
\usepackage{algorithmicx}  
\usepackage{algpseudocode}

\usepackage{caption}
\usepackage{float} 
\usepackage{subcaption}
\usepackage{tcolorbox}
\usepackage{diagbox}
\usepackage{multirow}
\usepackage{lipsum}
\usepackage{makecell}
\usepackage{pifont}
\usepackage{engord}
\usepackage{microtype}
\usepackage{amsmath}
\usepackage{booktabs}
\usepackage[table]{xcolor}
\usepackage{bm}
\usepackage{xcolor}
\usepackage{wrapfig}
\usepackage{graphicx} 
\usepackage{placeins}
\usepackage{dblfloatfix}

\newcommand{\dataset}{{AgentFail}}

\begin{document}

\title{Demystifying the Lifecycle of Failures in Platform-Orchestrated Agentic Workflows}


\author{Xuyan Ma}
\orcid{0000-0002-8514-2336}
\authornote{Also With State Key Laboratory of Complex System Modeling and Simulation Technology, Beijing, China; \\ Science \& Technology on Integrated Information System Laboratory, Beijing, China; \\ University of Chinese Academy of Sciences, Beijing, China;\\ }
\affiliation{%
\institution{Institute of Software \\ Chinese Academy of Sciences}
\city{Beijing}
\country{China}
}
\email{maxuyan2021@iscas.ac.cn}

\author{Xiaofei Xie}
\orcid{0000-0002-1288-6502}
\affiliation{%
  \institution{Singapore Management University}
  \city{Singapore}
  \country{Singapore}
}
\email{xfxie@smu.edu.sg}

\author{Yawen Wang}
\orcid{0000-0003-2854-4889}
\authornotemark[1]
\authornote{Corresponding author}
\affiliation{%
\institution{Institute of Software \\ Chinese Academy of Sciences}
\city{Beijing}
\country{China}
}
\email{yawen2018@iscas.ac.cn}

\author{Junjie Wang}
\orcid{0000-0002-9941-6713}
\authornotemark[1]
\authornotemark[2]
\affiliation{%
\institution{Institute of Software \\ Chinese Academy of Sciences}
\city{Beijing}
\country{China}
}
\email{junjie@iscas.ac.cn}

\author{Boyu Wu}
\orcid{0009-0001-9285-3419}
\authornotemark[1]
\affiliation{%
\institution{Institute of Software \\ Chinese Academy of Sciences}
\city{Beijing}
\country{China}
}
\email{boyu_wu2021@163.com}

\author{Mingyang Li}
\orcid{0009-0008-7936-5593}
\authornotemark[1]
\affiliation{%
\institution{Institute of Software \\ Chinese Academy of Sciences}
\city{Beijing}
\country{China}
}
\email{mingyang2017@iscas.ac.cn}

\author{Qing Wang}
\orcid{0000-0002-2618-5694}
\authornotemark[1]
\authornotemark[2]
\affiliation{%
\institution{Institute of Software \\ Chinese Academy of Sciences}
\city{Beijing}
\country{China}
}
\email{wq@iscas.ac.cn}


\renewcommand{\shortauthors}{Trovato et al.}

\begin{abstract}
Agentic workflows built on low-code orchestration platforms enable rapid development of multi-agent systems, but they also introduce new and poorly understood failure modes that hinder reliability and maintainability. Unlike traditional software systems, failures in agentic workflows often propagate across heterogeneous nodes through natural-language interactions, tool invocations, and dynamic control logic, making failure attribution and repair particularly challenging.
In this paper, we present an empirical study of platform-orchestrated agentic workflows from a failure lifecycle perspective, with the goal of characterizing failure manifestations, identifying underlying root causes, and examining corresponding repair strategies. We present {\dataset}, a dataset of 307 real-world failure cases collected from two representative agentic workflow platforms. Based on this dataset, we analyze failure patterns, root causes, and repair difficulty for various failure root causes and nodes in the workflow. Our findings reveal key failure mechanisms in agentic workflows and provide actionable guidelines for reliable failure repair, and real-world agentic workflow design.
\end{abstract}

\ccsdesc[500]{Software and its engineering~Software testing and debugging}

\keywords{Agentic workflows,
Failure lifecycle,
Failure attribution,
Root cause analysis,
Failure repair
}

\received{20 February 2007}
\received[revised]{12 March 2009}
\received[accepted]{5 June 2009}

\maketitle

\section{Introduction}
\label{sec:intro}

Large Language Models (LLMs) have recently shown remarkable capabilities in reasoning, planning, and knowledge-intensive tasks\cite{matarazzo2025surveylargelanguagemodels}, which promotes their widespread adoption across diverse application domains \cite{chen2023agentverse}.
Building on these advances, agentic systems
powered by LLMs are gaining increasing attention, as they enable multiple specialized agents to collaborate toward complex goals.
Such agentic systems have been applied to software development, information retrieval and research assistance \cite{guo2024largelanguagemodelbased}, where the coordination of agents often outperforms single-agent solutions.

\begin{figure*}[htbp]
    \centering
    \includegraphics[width=0.99\textwidth]{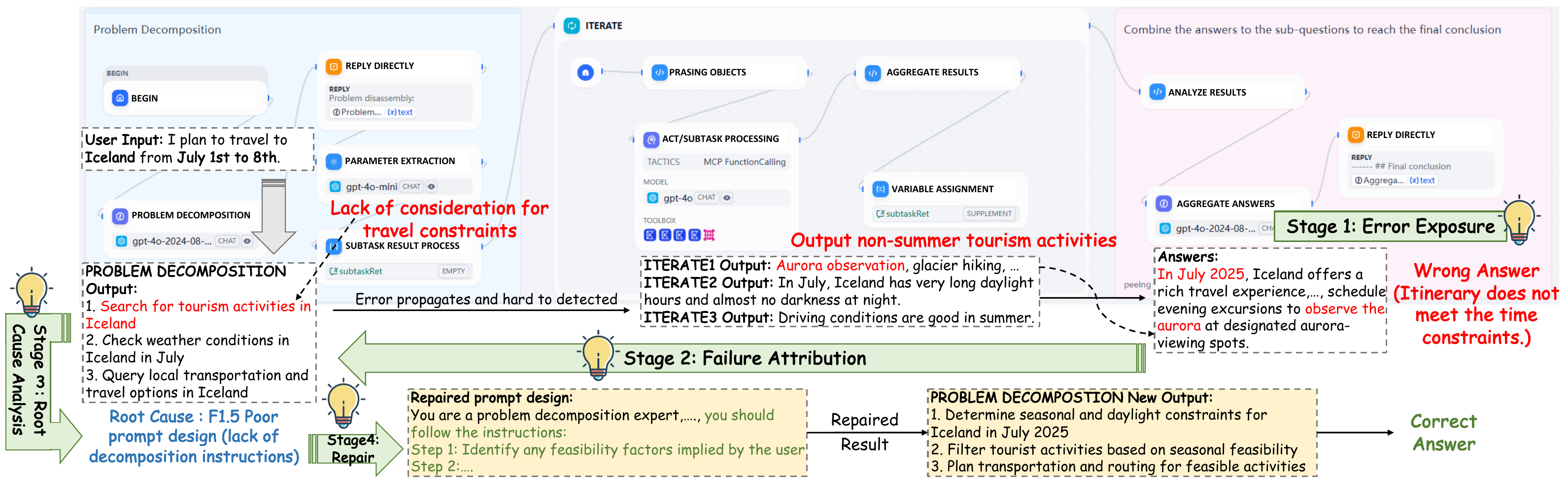}
    \caption{Example of failure lifecycle in agentic workflow.
    }
    \label{fig:intro}
    \end{figure*} 


To further lower the barrier to building such systems, a new wave of low-code agentic AI development platforms, such as Dify \cite{dify}, Coze \cite{coze}, n8n \cite{n8n} and AutoGen \cite{autogen}, has emerged.
These platforms provide intuitive workflow editors, pre-configured tool integrations, and flexible orchestration mechanisms, allowing users to rapidly prototype and deploy multi-agent solutions without extensive programming expertise. 
{Although these low-code and template-based tools have significantly lowered the bar for building agentic workflows, the inherent fragility of LLMs, such as their susceptibility to hallucinations, combined with the structural complexity of agentic orchestration makes such systems prone to failure in practice.}

{Similar to traditional software, repairing such failures to restore correct system behavior also requires attributing the failure, identifying its root cause, and then deriving a corresponding fix, which is a non-trivial process \cite{7390282}.
More than that, repairing such agentic workflows is more challenging. 
Unlike traditional software, which follows deterministic execution logic and well-defined control flows, real-world agentic workflows are driven by large language models and dynamic interactions. As a result, failure traces frequently exhibit rich and entangled interaction information, including natural language prompts, multi-step agent communications, and tool-invocation records \cite{Epperson_2025}. 
Due to the inherent ambiguity of natural‑language descriptions, attributing the root cause of a failure becomes highly complex. 
Furthermore, fixing these failures does not merely involve code modifications; it may also require refining prompts, restructuring the workflow, and other adjustments.
Figure \ref{fig:intro} illustrates an example of this process. At the final result presentation stage, the user observes that the model outputs a travel plan that  does not meet the given time constraints. 
By carefully tracing back through 7 intricate node dependencies, systematically examining the conversational context, tool-call sequences, and natural-language reasoning chains of each agent, and cross-referencing their inputs/outputs with the intended business logic, the user eventually attributes the error: the planning agent made a mistake during task decomposition because its prompt design did not explicitly provide decomposition constraints, causing it to overlook the travel time in task decomposition.
Finally, the user revises the prompt to instruct the LLM to identify any feasibility factors implied by the user, thereby fixing the issue.}

{Several existing studies have begun to explore failure localization and repair in LLM‑based multi‑agent systems. FAMAS \cite{ge2025} attempted to identify responsible agents and steps using spectrum analysis. AgenTracer \cite{zhang2025agentracer} constructed TracerTraj via programmatic fault injection and counterfactual corrections, then trained a dedicated failure attributor through reinforcement learning with multi‑grained rewards. 
Song et al. \cite{song2025aegis} designed Aegis, which employs a set of targeted environment optimizations for failure repair. 
Wang et al. \cite{wang2025maestro} introduced Maestro, a framework‑agnostic holistic optimizer that jointly searches over graphs and configurations to maximize agent quality under explicit rollout/token budgets. 
However, these localization and repair attempts are generally optimization‑based and do not fundamentally understand the failure mechanisms or corresponding repair strategies inherent in this new paradigm of agent systems.}

{Cemri et al. \cite{cemri2025multi} proposed MASFT (Multi‑Agent System Failure Taxonomy), which identifies 14 unique failure modes involving specification issues, inter‑agent misalignment, and task verification. 
Lu et al. \cite{lu2025exploringautonomousagentscloser} analyzed agent failures from the perspective of execution stages: task planning, task execution, and response generation. 
Zhu et al. \cite{zhu2025llm} proposed AgentErrorTaxonomy, a failure modes spanning memory, reflection, planning, action, and system-level operations.
Yet, these works largely describe failure manifestations without delving into their underlying root causes, let alone proposing concrete repair strategies. 
For example, in the MASFT taxonomy \cite{cemri2025multi}, the most frequent failure is ``step repetition'', which is merely a surface‑level symptom. In Lu et al.\cite{lu2025exploringautonomousagentscloser}, we can know the manifestations of failure ``Infinite loop with same response'', but still find it difficult to grasp the underlying reasons for the failure.
Even when developers are informed of this failure, they cannot link it back to a root cause, which leaves them without actionable guidance for repair.  
This lack of holistic understanding fundamentally hinders progress in achieving reliable workflow repair.
}
\begin{table*}[!bp]
\centering
\caption{{Dataset Information. For test data sources with a large scale, we randomly select a subset of 100 samples to form the test set. We use the format \texttt{Hand-crafted-Platform Name-Task Name} to name the test data created by ourselves. The failure logs come from our run or community.} 
}
\label{tab:dataset}
\resizebox{0.92\textwidth}{!}{
\begin{tabular}{ccccc|c}
\toprule
\rowcolor{gray!15} \textbf{Platform} & \textbf{Task} & \textbf{Structure} & \textbf{Test Data Source} & \textbf{Test Data size} & \textbf{\makecell[c]{Failure Log \\ (Run + Community)}} \\ 
\midrule
\multirow{5}{*}{Dify} 
& Code Generation & Looping & HumanEval \cite{humaneval} & 163 & 23 (21+2) \\
& Program Repair & Looping & SWE-bench \cite{jimenez2023swe} & 100 & 62 (62+0) \\ 
& Product QA Assistant & Branching & Hand-crafted-Dify-QA & 100 & 18 (17+1) \\
& Travel Assistant & Parallel & TravelPlanner \cite{travelplanner} & 180 & 32 (31+1) \\
& Deep Research & Hybrid & ninja-x-deepreasearch \cite{ninjax_deepresearch} & 182 & 27 (26+1) \\
\midrule
\multirow{5}{*}{Coze}
& Product QA Assistant & Hybrid & {Hand-crafted}-Coze-QA & 100 & 37 (34+3) \\ 
& Travel Assistant & Parallel & TravelPlanner & 180 & 31 (28+3) \\
& Market Research Assistant & Serial & Hand-crafted-Dify-Market & 100 & 30 (29+1) \\
& Deep Research & Serial & ninja-x-deepreasearch & 182 & 27 (25+2) \\
& Industry Analysis & Parallel & Hand-crafted-Dify-Industry & 100 & 20 (20+0) \\ 
\midrule
\rowcolor{gray!15} \textbf{Sum} & & & & \textbf{1,387} & \textbf{307 (293+14)} \\
\bottomrule
\end{tabular}}
\end{table*}
{To bridge this gap, we present an empirical study from the perspective of the failure lifecycle in platform‑orchestrated agentic workflows, aiming to characterize failure manifestations, identify potential root causes, and explore corresponding repair strategies. 
We first introduce a dataset named {\dataset}\footnote{Website URL: https://github.com/Jenna-Ma/JaWs-AgentFail} that explicitly captures the entire failure lifecycle to support systematic analysis. 
It consists of 307 failure trajectories collected from two representative agentic workflow development platforms, annotated through multi‑round expert labeling with reliable consensus. 
Each instance includes the user query, the agentic workflow configuration, execution failure logs, and detailed lifecycle annotations covering failure location, root cause, and repair strategy. 
Based on this dataset, we conduct detailed qualitative analysis, develop automated technique for repair, and evaluate the performance of different failure types in terms of repair success rate and cost.
Moreover, we provide in‑depth examination of failure occurrence patterns, root causes, and repair difficulty across various types of {failure root causes and }agent nodes. Together, these analyses offer a clearer understanding of the failure mechanisms in this emerging paradigm of agentic systems and deliver actionable guidelines to facilitate the development of more reliable and robust automated failure attribution and repair approaches.}
Besides, we also provide practical guidelines for developers to construct more robust agentic workflows on platform,
ensuring that our study not only deepens understanding but also supports real-world platform practices.

\section{Background}
\label{sec:background}

\textbf{Agentic AI Development Platform.}
Modern agentic AI development platforms (e.g., Dify, Coze, n8n) expose agentic workflow construction through a node-based paradigm, where each node corresponds to a functional unit. By connecting nodes, users can define complex multi-agent system that combine reasoning, control, and external interactions.
The \textbf{node types} are summarized as follows:
1) Start and Termination Nodes;
2) LLM and Agent Nodes;
3) Knowledge Nodes;
4) Logic and Control Nodes;
5) Code and Template Nodes;
6) Tool and Integration Nodes.
These nodes extend the platform’s capability beyond built-in components, enabling workflows to interact with external services, databases, or custom tools. These node types are further analyzed in Sec. \ref{subsec:failure-attribution} except Start and Termination nodes because they serve as the entry and exit markers of the workflow and do not directly contribute to execution failures.

\textbf{{Failure Definition.}}
{Since we would like to collect failure trajectories of agentic workflows, we first need to determine whether the current execution has deviated from expected behavior. In our setting, the execution of system interrupts is directly regarded as a failure. For the concluded executions, we adopt the following two evaluation strategies for different types of tasks:} 

{\textit{Ground Truth Comparison}}: For tasks with deterministic ground truth or validation mechanism (e.g., code generation), we directly compare the outputs against the provided ground truth. One execution is considered successful if the output either passes test-based verification (e.g., functional correctness in code execution) or exactly matches the reference solution.

{\textit{Multi-LLM Judge}}: For tasks lacking deterministic ground truth (e.g., task planning, deep research), we adopt LLM-as-a-judge \cite{gu2024survey} technique. Specifically, we use multiple independent LLMs as evaluators, each tasked with assessing whether the system output satisfied the task requirements. Then, a consensus voting strategy is applied, where an output was marked as correct if a majority of LLM judges agreed on its validity.

    



To identify the root location of failure where the failure originates,
we consider node actions that could be decisive: 
if replacing a wrong action with a correct one (while keeping previous steps unchanged) changes the trace from failure to success, we regard this action as decisive.
When multiple such actions exist, we follow an earliest-in-time principle and select the first decisive action as the failure root.
Based on the identified location, we further analyze and summarize the underlying root causes of failure.

\section{The {\dataset} Dataset}
\label{sec:dataset}

We collect systems execution data from two popular agentic AI development platforms, Dify and Coze, which provide the capabilities of visual workflow composition and tool integration. 
The dataset {\dataset} contains 307 failure logs from ten platform-orchestrated agentic workflows in total, {with} five from each platform.
Each instance in the dataset includes four elements: (1) Query, a query obtained from the real test case; (2) Failure log: the full conversational trace of a system failing to complete the task. (3) Workflow and configuration: including the node orchestration structure and the information of each node, like agent's name, agent's prompt, code, tool configuration. 
(4) Annotations: the root location of failure, the root cause taxonomy labels, further explanation about  why the failure took place and the corresponding repair strategy.

\subsection{System selection}
The agentic workflows span a variety of task categories, including software development, information insight, task planning, and question \& answering. 
Both platform-provided templates and user-defined systems we select from the platforms are included to ensure coverage of typical cases. 
To ensure diversity, we consider not only the task types but also the structural design of systems, 
covering five common categories: serial, parallel, branching, looping, and hybrid structures, whose details can be found on the website. 
This choice enables our analysis to cover diversified failure modes that are specific to both task semantics and orchestration patterns.

\subsection{Failure log collection}
Failure logs analyzed in this study are obtained from two sources: (1) open-source community contributions, where users had publicly shared, whose example can be found on the website.
From these reports, we reproduce the executions based on the user input to construct failure cases;
and (2) our own controlled runs, in which we execute public datasets {(e.g., HumanEval \cite{humaneval}, TravelPlanner \cite{travelplanner})} or hand-crafted datasets based on the task, 1,387 test data in total, 
on corresponding workflow and systematically record the traces. {For hand-crafted datasets {(denoted as Hand-crafted-Platform name-Task name)}, 
we carefully design the construction criteria to ensure the representativeness and diversity of the inputs, which can be found on our website. 
For each run, success or failure is determined by the criteria described in Sec. \ref{sec:background}.
Finally, we collect 307 failure logs totally and these combined sources ensure that our dataset captured both naturally occurring failures and reproducible benchmark failures.
{Detailed stats of {\dataset} are listed in Table \ref{tab:dataset}.}

\subsection{Failure Root Location \& Cause Annotation}
\label{subsec:failure}
{To systematically identify both where failures originate and why they occur, we identify the failure root location and root cause based on multiple rounds of annotation.}
{During the process,} we adopt Grounded Theory (GT) annotation method \cite{1968The}, which is a qualitative research method that directly constructs theories from empirical data, to identify root cause patterns.

(i) \textbf{Independent Annotation}. 
Three annotators with expertise in agentic systems independently examined the execution traces, where
each annotator manually identifies the decisive errors and corresponding root cause based on the expert knowledge and understanding of failure logs.
Additionally, each expert is instructed to separate their annotations into two categories: 
cases where they are fully confident in the identified failure, and cases involving any degree of uncertainty.
(ii) \textbf{Consensus Building}. Annotators then focus specifically on the cases marked as uncertain in the first step. These instances are jointly reviewed, and through collaborative discussion, agreement is reached on the final labels.
(iii) \textbf{Cross Validation}. Finally, we adopt a cross-validation procedure. Each expert reviews the annotations made by others to assess the consistency of labeling standards. If any discrepancies are identified, the annotators engage in further discussion and, when necessary, re-annotate the data together until consensus is achieved. By incorporating multiple perspectives and enforcing agreement among annotators, this process enhances the reliability of the final annotations.

\section{Demystifying Failure Lifecycle}
\label{sec:result}



\subsection{Failure Attribution}
\label{subsec:failure-attribution}

\subsubsection{\textbf{Failure Manifestation}}
\label{subsubsec:statistical}
We conduct a comprehensive analysis of all failure attribution nodes in the dataset. Specifically, we first examine the types of nodes that are identified as being responsible for the final failure, as defined in Sec. \ref{sec:background}.
Secondly, we also analyze their positions within the workflow execution, including the distance to the point where failures are ultimately exposed and their distances relative to the start and end of the workflow.

\textbf{Node Type.} {We present the types of nodes that are more likely to introduce failures in {\dataset} to demystify the failure distribution. }
As Figure \ref{fig:pie}(a) shows, LLM \& Agent nodes dominate failure root causes, accounting for over half of all cases, indicating that agent-level reasoning and planning are the primary sources of failures in agentic workflows. 
The logic \& control nodes and knowledge nodes are the second and third most error-prone nodes respectively, suggesting that unreasonable orchestration and wrong knowledge retrieval also play significant roles. 
{Code and template nodes, whose failures are related to traditional software, account for only a very small proportion of all failures cases, highlighting that in this emerging agent-based systems, the failure patterns and corresponding repair priorities are fundamentally different from those in traditional software systems.}
\textbf{Position.} 
We further analyze the failure root position by measuring the distance $\mathbb{D}$ between the root cause node and the failure exposed node, to characterize how long an error propagates through the workflow before becoming observable.
Since agentic workflows vary in length, we normalize this distance $\mathbb{D}$ by dividing its value by the total length of the workflow, enabling fair comparison across workflows of different scales.
\begin{figure}[bp] 
    \centering
    \includegraphics[width=0.49\textwidth]{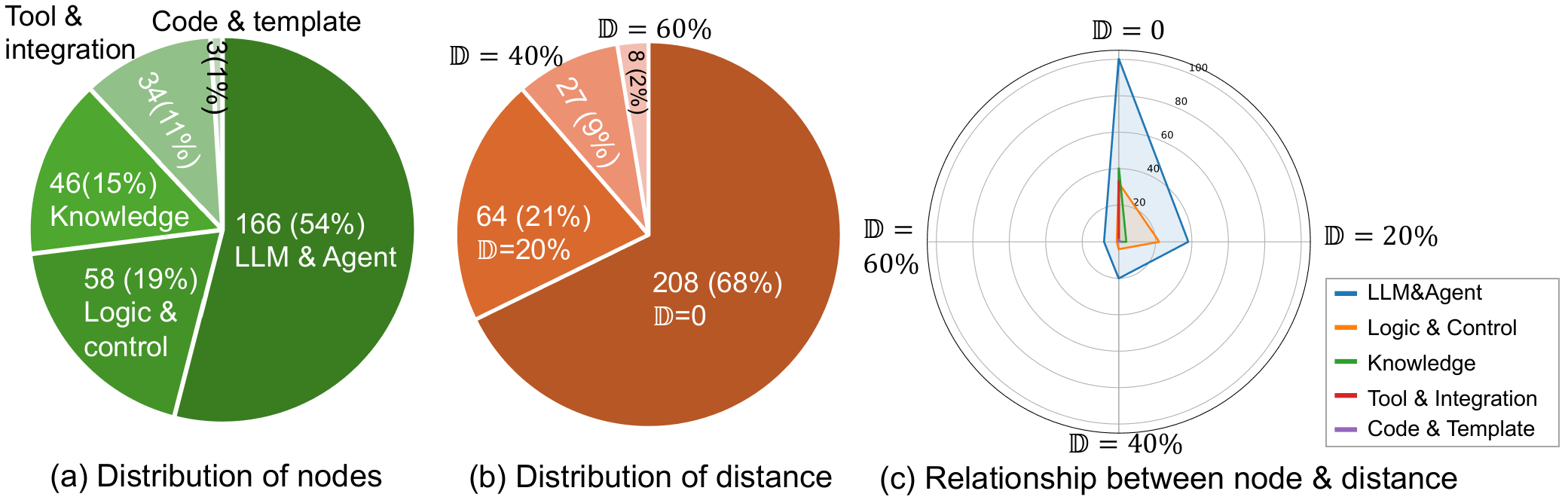}
    \caption{Statistic of Distribution. In (c), each concentric ring indicates the number of failure cases.} 
    \label{fig:pie}
\end{figure}

    


Figure \ref{fig:pie}(b) illustrates the distribution of the normalized distance across all failed workflows. 
{In 32\% of failures, the root cause node differs from the failure symptom node, with propagation distances exceeding 40\% of the workflow length in over 10\% of cases.
Particularly, nodes such as LLM\&Agent and logic \& control are prone to such long‑distance error propagation, significantly amplifying the challenge of failure repair. As Figure \ref{fig:pie}(c) shows, for LLM \& Agent nodes, approximately 40\% of failure root causes occur at locations different from where the failures are exposed.
This proportion is even higher for Logic \& Control nodes, reaching 45\%, indicating that failures originating from these nodes are more likely to propagate across multiple nodes and manifest as non-local failures. }
Due to the space limitation, we present the remaining analysis result on the website, including the distance between root cause nodes and the start node, as well the termination node. These analyses help characterize how early failures are introduced during workflow execution and to what extent their effects propagate, providing further insights into the non-local nature of failures in agentic workflows.

\subsubsection{\textbf{Automatic Failure Attribution Performance}}
\label{subsubsec:attribution-bench}
\begin{figure}[tbp]
    \centering
    \includegraphics[width=0.49\textwidth]{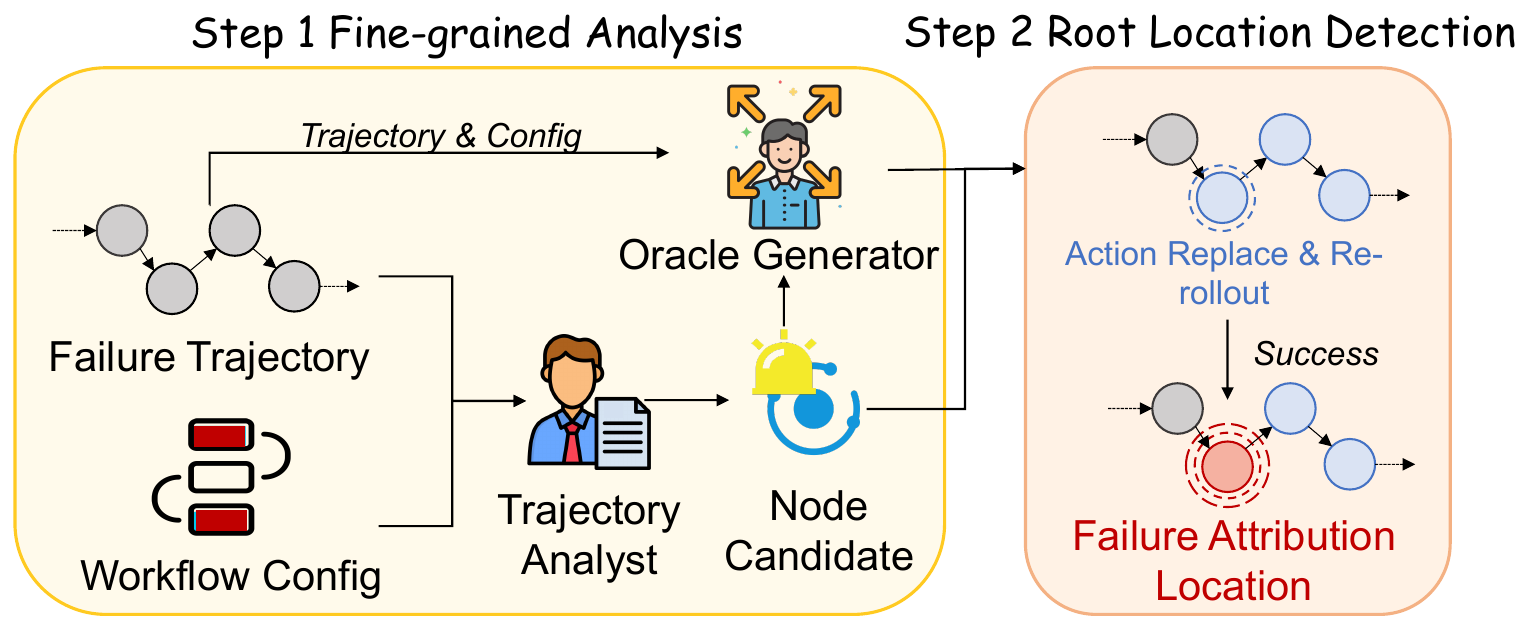}
    \caption{The process of failure attribution.
    } 
    \label{fig:attribution}
    \end{figure} 

\begin{figure*}[tbp]
    \centering
    \includegraphics[width=0.99\textwidth]{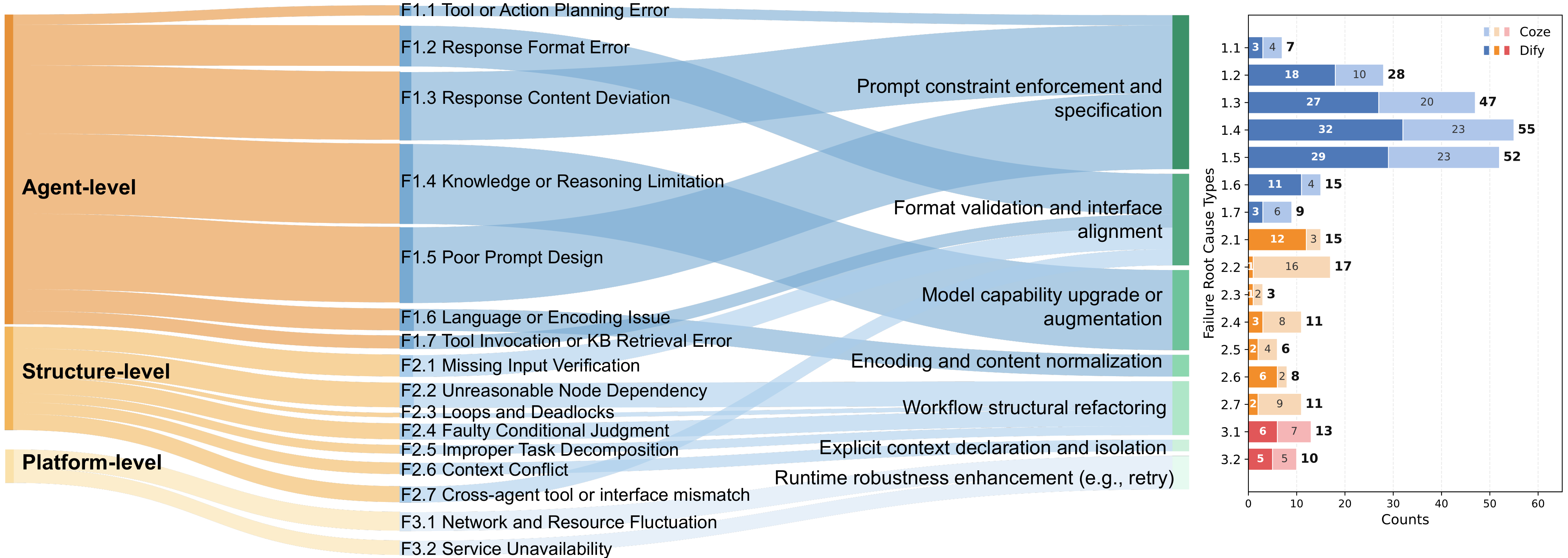}
    \caption{Failure Root Cause and Repair Strategy Taxonomy. 
    }
    \label{fig:taxonomy}
    \end{figure*}



As mentioned in Sec.\ref{sec:intro}, failure attribution is challenging and often requires substantial manual effort. {Therefore, we aim to investigate the performance of automated failure attribution for agentic workflows. Following prior work \cite{zhu2025llm}, we customize an automated framework for failure attribution tailored to agentic workflows, as Figure \ref{fig:attribution} shows. It consists of two key stages: in Step 1 (Fine-grained Analysis), we jointly analyze the failure trajectory and the workflow configuration to systematically examine agent interactions, control flows of workflow, and intermediate outputs, and identify a small set of candidate suspicious nodes that are likely to be responsible for the observed failure. This effectively narrowed down the search scope for failure attribution.
In Step 2 (Root Location Detection), we perform re-rollout-based counterfactual verification on the candidate nodes identified in Step 1. Specifically, we selectively replace the outputs of a candidate node with corrected ones (generated by Oracle Generator) while keeping the rest of the execution unchanged, and re-execute the workflow to observe whether the failure is resolved. The earliest node whose correction turns the failure into a success is identified as the root failure location. Together, these two steps transform failure attribution from heuristic inspection into a principled, causality-oriented process.}

\begin{table*}[hbtp]
\centering
\caption{Performance comparison of failure attribution, root cause identification. 
RLDetect represents the root location detection in Figure \ref{fig:attribution}.
LO and TA respectively represent failure location and root cause taxonomy.
}
\label{tab:benchmark}
\resizebox{0.82\textwidth}{!}{
\begin{tabular}{cc|cccc}
\toprule
 \multicolumn{2}{c|}{\textbf{Failure Attribution}} 
& \multicolumn{4}{c}{\textbf{Root Cause Identification}} \\
\cmidrule(lr){1-2}
\cmidrule(lr){3-6}
w/o RLDetect
& with RLDetect
& w/o LO, w/o TA 
& w/o LO, with TA 
& with LO, w/o TA 
& with LO, with TA \\
\midrule
46.3\% & 65.8\% & 9.6\%  & 27.4\% & 22.8\% & 45.6\% \\
\bottomrule
\end{tabular}}
\end{table*}

{As shown in Table \ref{tab:benchmark}, 
failure attribution without re-rollout verification (i.e., w/o RLDetect) achieves 46.3\% accuracy.
In contrast, incorporating the re-rollout-based root location detection (illustrated in Figure \ref{fig:attribution}) improves the attribution accuracy to 65.8\%.
The result indicates that failures in 
agentic workflows often involve cross-node dependencies and long-range failure propagation, which makes one-shot, end-to-end attribution vulnerable to being misled by local evidence, noisy execution logs, or surface-level symptoms.
Without explicit validation, attribution models may incorrectly assign responsibility to nodes that are temporally or structurally close to the error manifestation but not causally responsible.
}

\subsection{Failure Root Cause}
\label{subsec:root_cause}

\subsubsection{\textbf{Taxonomy \& Distribution}}
We systematically analyze the failure cases observed in agentic workflows and construct a structured,
repair-oriented failure root cause taxonomy, as shown in Figure \ref{fig:taxonomy}. 
The leftmost layer classifies failures according to the system abstraction level at which they originate. The middle layer specifies concrete failure root causes, while the rightmost layer summarizes corresponding repair strategies.
Agent-level failures capture failures that occur within a single agent, primarily due to limitations of the underlying LLM or its interaction with local resources; 
Structure-level failures arise from coordination or communication among other agents, often linked to workflow orchestration structures; 
Platform-level failures are attributed to the underlying platform or runtime environment.
{As mentioned in Sec.\ref{sec:intro}, many existing taxonomies primarily characterize failure manifestations, such as ``step repetition'' in MASFT\cite{cemri2025multi} and ``infinite loop with same response'' in Lu et al. \cite{lu2025exploringautonomousagentscloser}, which neither aid in understanding the root causes of failures nor offer guidance for repair. 
However, our taxonomy is repair‑oriented: each category corresponds to a fundamental cause of failure and is mapped to concrete repair strategies, thereby directly supporting actionable fixes.}
{Figure~\ref{fig:taxonomy} also demonstrates the occurrences of each root cause.}
Overall, agent-level failures dominate the dataset, with knowledge and reasoning limitations (F1.4) and poor prompt design (F1.5) being the most frequent categories. 
Structure-level failures are less common but remain important, with missing input validation (F2.1) and unreasonable node dependency (F2.2) appearing most often, while structural issues such as loops, deadlocks (F2.3), and improper task decomposition (F2.5) are relatively rare.
Platform-level failures constitute the smallest proportion, including network or resource fluctuations (F3.1) and service unavailability (F3.2). Due to space limitations, additional data analyses are provided on our website.

\subsubsection{\textbf{Automatic Root Cause Identification Performance}}
{We further explore how automated techniques can identify the root cause, and} investigate whether failure attribution and our proposed root cause taxonomy can assist this process. {Specifically, we prompt the LLM to act as a root cause analyst. Given the failure attribution node, together with the execution trajectory and workflow configuration, we leverage our proposed root cause taxonomy to guide the analyst toward structured and systematic reasoning about the failure.
In our experiments, we consider four different settings by varying whether the failure attribution location and the root cause taxonomy are provided.}
Table~\ref{tab:benchmark} shows that providing both root cause taxonomy and failure location substantially facilitates root cause identification.
When neither taxonomy nor failure location is available, the model achieves only 9.6\% accuracy, indicating that root cause identification in a fully unguided setting is extremely challenging. 
Introducing the root cause taxonomy alone leads to a significant performance improvement, boosting accuracy to 27.4\%. This improvement suggests that taxonomy provides strong conceptual guidance by constraining the model’s reasoning space to semantically meaningful failure categories, enabling more structured causal inference even without explicit localization information.
However, the gains from location alone are consistently smaller than those achieved by taxonomy alone. 
This highlights that root cause identification is not merely a localization problem, but fundamentally a semantic reasoning task. 
While failure location answers where the failure is observed, it does not directly explain why the failure occurs, leaving substantial ambiguity in complex agentic workflows.

The best performance is achieved when both taxonomy and failure location are provided simultaneously, with accuracy reaching 45.6\%. This synergy indicates that taxonomy and localization play complementary roles. Taxonomy offers an abstract causal framework that defines the space of plausible failure mechanisms, while failure location grounds this reasoning in concrete execution traces. Together, they jointly constrain both the semantic and structural dimensions of the problem, enabling more precise and stable root cause identification. Meanwhile, the 45.6\% accuracy rate also indicates that identifying the root cause is a very challenging task. 

\begin{figure}[bp]
\vspace{-0.02in}
    \centering
    \includegraphics[width=0.5\textwidth]{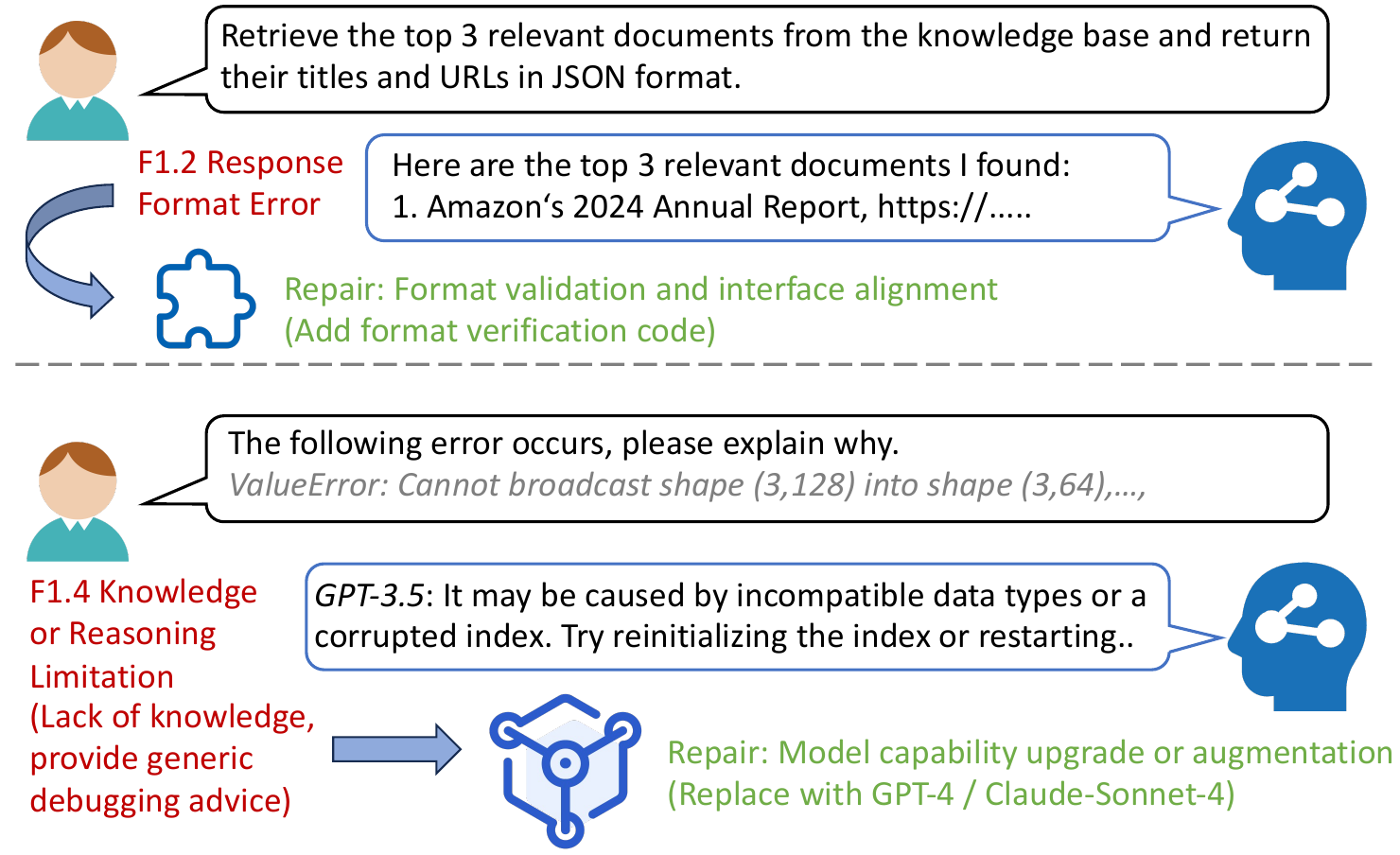}
    \caption{Examples of failure and repair.}
    \label{fig:repair-example}
    \end{figure}

\subsection{Failure Repair}
\subsubsection{\textbf{Repair Strategy Taxonomy}}

We systematically study 
what are the repair strategies for repairing failed agentic workflows.
By analyzing how specific failure causes manifest in agentic workflows, we identify the actionable strategies that directly target the underlying issues, as shown in Figure \ref{fig:taxonomy}. {Figure \ref{fig:repair-example} illustrates two failure modes and the corresponding repair method in agentic workflows: response format errors (F1.2) and knowledge or reasoning limitations (F1.4). The first failure arises from misaligned output formats, which can be repaired via adding format verification module; the second stems from insufficient model capability, requiring upgrades or external augmentation for effective comprehension and debugging.}
\subsubsection{\textbf{Automatic Repair Performance}}
\label{subsec:repair}


We construct a repair framework based on popular agent-based methods in traditional program repair domain. Drawing inspiration from established methods such as SWE-agent \cite{yang2024swe} and Agentless \cite{xia2024agentless}, we construct a failure repair team consisting of four specialized experts, who collaboratively perform the repair task in a staged manner, with each expert responsible for a distinct phase of the repair process.
During this process, the experts leverage failure trajectories, workflow configurations, as well as explicit information about failure locations and root causes {(if any)} to guide their decisions and actions. This setting enables us to examine how failure attribution and root cause knowledge contribute to effective and reliable workflow repair. 
Accordingly, we apply {two other} experimental settings based on whether to use failure attribution and root cause information 
to explore the significance of the two components. Notably, the setting ``w/o LO \& with RC'' is not included since it is an unreasonable experimental setup.
\begin{figure}[tbp]
    \centering
    \includegraphics[width=0.49\textwidth]{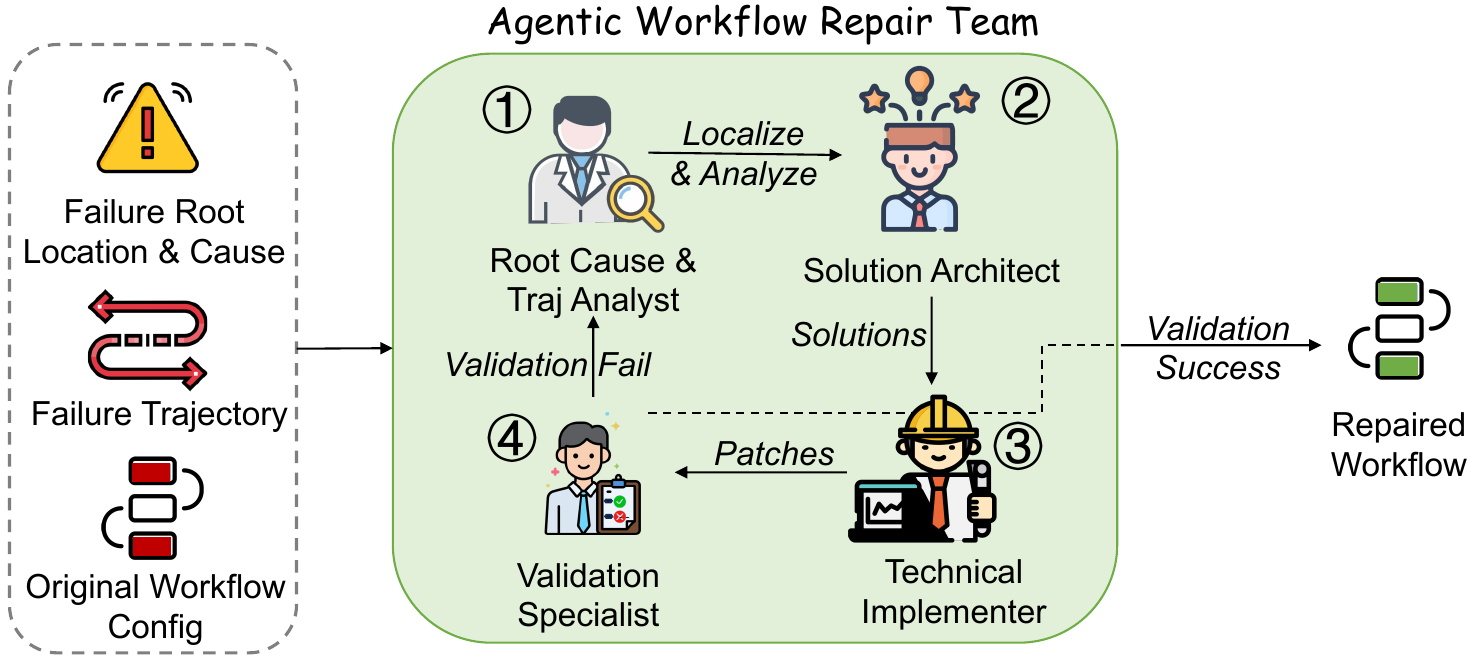}
    \caption{The process of repair.}
    \label{fig:repair}
    \end{figure}


\begin{table}[bp]
\centering
\caption{Performance comparison on failure repair. LO and RC respectively represent the failure location and root cause.}
\label{tab:repair-result}
\resizebox{0.48\textwidth}{!}{
\begin{tabular}{l|cc}
\toprule
\textbf{Methods}  & \textbf{Repair Success Rate} & \textbf{New Failure Rate}  \\
\midrule
w/o LO \& w/o RC & 16.3\% & 13.9\%\\
with LO \& w/o RC & 34.9\%& 8.4\%\\ 
\rowcolor{gray!15} with LO \& with RC & \textbf{66.8\%} & \textbf{2.8\%}\\
\bottomrule
\end{tabular}}
\end{table}


\begin{figure*}[htbp]
    \centering
    \includegraphics[width=0.99\textwidth]{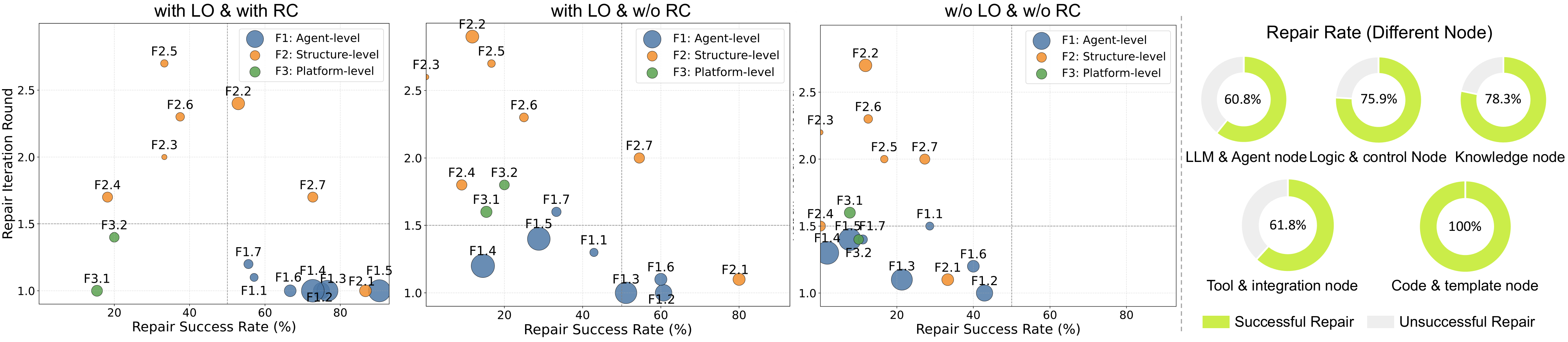}
    \caption{Repair Difficulty of Different Root Cause and the repair rate of different node. (In first three subfigures, the size of each point represents the number of failures labeled as that root cause.)}
    \label{fig:repair-four}
    \end{figure*}

{Firstly, we investigate the \textbf{repair success rate and whether the act of repairing introduces any new failures}. The new failure rate is measured by re-running the repaired workflow on the original set of test inputs that previously passed ($1387 - 307$) and computing the proportion of failed cases.
As shown in Table \ref{tab:repair-result}, as the repair guidance information gradually becomes more comprehensive, the repair success rate significantly increases and the new error rate continuously decreases. Especially when both the failure root location and the root cause information are provided simultaneously, the repair success rate reaches 66.8\%, and the new error rate is only 2.8\%, demonstrating a more effective and safer repair outcome. When neither the failure location information nor the root cause information is provided, a large number of new errors are introduced, indicating that the repair behavior is unreliable. }

Secondly, we analyze \textbf{repair performance across different root cause categories} by measuring repair success rates and the number of repair iterations.
As shown in Figure \ref{fig:repair-four}, root causes located closer to the lower right region of the figure correspond to higher repair success rates and lower repair costs, and are therefore easier to repair.
We can find that agent-level root causes 
are clustered in the region with high repair success rates and low iteration costs, indicating that they are relatively easy to repair through targeted modifications.
In contrast, structure-level root causes, particularly those involving control logic, node dependencies, or task decomposition, exhibit lower repair success rates and require more repair iterations. F2.1 (Missing input verification) is an exception: although it is related to workflow orchestration, it can often be repaired by simply adding input validation mechanisms, resulting in a relatively high repair success rate.
Moreover, we analyze the sensitivity of repairing different root causes to the availability of failure localization and root cause information.
We find that the causes related to formatting (F1.2), encoding (F1.6) have a relatively low sensitivity to the failure location and root cause information. Even in the absence of clear guidance, they can often be repaired. 
{We also observe that underlying causes that require deeper semantic understanding (e.g., F1.5 Poor Prompt Design) exhibit a substantial decrease in repair performance when root cause information is absent. Moreover, when the failure root location is further removed, the repair rate continues to drop. 
}

{Furthermore, we investigate the \textbf{repair rates across different node types}, revealing a clear relationship between node characteristics and repair effectiveness, as shown in Figure \ref{fig:repair-four}. 
We can see that nodes with more deterministic structures, such as code and template nodes, exhibit consistently higher repair success, whereas LLM \& Agent nodes and tool \& integration nodes remain more difficult to repair. The observed differences in repair rates suggest that workflow developers should explicitly account for node heterogeneity during design. In particular, failures originating from LLM \& Agent nodes and tool and integration nodes are harder to repair, indicating that these nodes should be treated as high-risk components. 
For more deterministic components, such as code, template, and knowledge nodes, developers can rely on stricter specifications and automated checks, as errors in these nodes are more likely to be localized and fully repairable. }

\section{Discussion}
\label{sec:discussion}



In order to design more effective automatic failure repair methods, we further analyze the cases that are not repaired and identify three main reasons: (1) the bottleneck of the repair agent's capabilities, making it difficult to obtain reliable correct answers, which often leads to incorrect repair directions; (2) the inability to break out of cognitive boundaries, which result in very few or no attempts at repair; (3) platform mechanism restrictions, which make it difficult for ordinary developers to avoid current failures.
Based on these observations, we derive the following recommendations for future automated failure repair systems. (i) First, for collaborative repair teams, it is encouraged for them to \textbf{actively challenge existing assumptions and explore different viewpoints}. In this way, they might achieve this goal by constructing broader or opposing test scenarios. This helps to overcome cognitive biases and enables the repair work to go beyond the current reasoning scope of the team. (ii) Secondly, \textbf{incorporating memory mechanisms or template-based repair strategies} can be beneficial, as repair agents may have encountered similar workflow failures in the past, yet such experiences are often underutilized, preventing effective knowledge transfer across repair attempts. 

Our ultimate goal is to enable developers to create more robust agentic workflows, so
we summarize the following actionable guidelines to help improve the process of developing workflows:
(i) \textbf{Clear role specification and modular prompt design} can mitigate planning errors and response misalignments, two of the most frequent failure root cause categories we observed (related with F1.3 Response content deviation, F1.5 Poor Prompt design).
(ii) {\textbf{Explicit input and output validation}} should be incorporated into nodes to prevent cascading errors from malformed data (related with F1.2 Response formatting error, F2.1 Missing input validation). 
(iii) \textbf{Comprehensive checks or fallback mechanisms}, such as secondary validation agents or alternative tool paths, help solve the problems at the local level before they propagate across the workflow (related with F2.4 faulty conditional judgment). Sometimes, such comprehensive checks may introduce additional overhead, which requires a balance between robustness and efficiency.
(iv) \textbf{Progressive workflow design}, which starts with simple serial or parallel flows and gradually introduces complex patterns. It can help build more robust systems (related with F2.2 unreasonable node dependency, F2.3 loops and deadlock).

\section{Related Work}
\label{sec:rw}


\textbf{LLM Multi-Agent Systems.}
{Recent work has explored LLM-based MAS across a variety of application settings. A representative line of research focuses on agents for software engineering tasks, such as code generation, and program repair, exemplified by systems like AutoGen \cite{autogen}, ChatDev \cite{qian2024chatdev}, MetaGPT\cite{metagpt}, AutoAgents\cite{chen2024autoagents} and SWE-agent \cite{yang2024swe}. 
For example, MetaGPT \cite{metagpt} organizes LLM agents into structured roles that follow software engineering workflows to collaboratively generate complex software artifacts from high-level requirements. SWE-agent \cite{yang2024swe} focuses on autonomously program repair through iterative interaction with code repositories, tools, and execution environments.
In addition, there are also some agents designed for other types of tasks. 
Both Magentic-One\cite{Magentic} and Captain Agent\cite{song2025captain} are capable of executing tasks from the GAIA\cite{gaia} benchmark.
Magentic-One employs a multi-agent system to collaboratively solve GAIA tasks, while Captain Agent leverages an automatically constructed agent team to perform task decomposition and execution, enabling effective problem solving across diverse GAIA scenarios. CAMEL\cite{li2023camel} introduces a role-playing communication framework that guides chat-based language agents to cooperate autonomously via prompt-driven interactions. 
The agentic workflows studied in this paper are built on low-code platforms. However, they form a distinct subclass characterized by visual, template-driven configurations rather than pure code, which introduces unique challenges in workflow design, debugging, and repair.
}
\textbf{{Failure Analysis for agentic systems.}}
{As LLM-based MASs grow increasingly complex, failures in agent coordination \cite{wang2025improving,marro2024scalable}, tool usage \cite{shen2024small}, and workflow execution have become a prominent challenge. Recent studies have begun to explore failure localization, diagnosis, and repair in such systems.
Some works focus on failure localization and responsibility attribution. 
Zhang et al. \cite{zhang2025agent} constructed Who\&When dataset to study agent- and step-level failure attribution, 
while Ge et al. \cite{ge2025} proposed methods to estimate agent responsibility in failed executions.
Zhang et al. \cite{zhang2025agentracer} investigated fine-grained failure localization using curated trajectories. 
Beyond failure localization, some efforts explored failure repair and optimization, such as Aegis \cite{song2025aegis} and Maestro \cite{wang2025maestro}, which aim to improve system robustness through automated optimization.
Another line of work seeks to analyze the root cause of agent failures. Cemri et al. \cite{cemri2025multi} proposed the MAST taxonomy to summarize common failure patterns in multi-agent systems. Lu et al. \cite{lu2025exploringautonomousagentscloser} and Zhu et al. \cite{zhu2025llm} analyzed failures from the perspectives of execution stages and agent capabilities, respectively.
However, existing studies primarily focus on where failures occur or how they manifest, rather than why they arise and how to repair them. Many identified failure types remain at the level of surface phenomena and do not directly translate into actionable repair guidance. Our work addresses this limitation by emphasizing root-cause–level analysis and proposing a structured taxonomy that bridges failure manifestations and systematic workflow repair.}

\section{Conclusion}

Unlike traditional software, failure traces in emerging agentic workflows contain complex interaction information, including natural language prompts, multi-agent communications, and tool invocation logs, making their repair inherently more challenging.
This paper presents a systematic investigation of the whole failure lifecycle in platform-orchestrated agentic workflows. Through the construction of AgentFail, a dataset of 307 annotated failure traces, we enable fine-grained analysis of where failures originate, how they propagate, and how to repair them. 
We introduce a repair-oriented root cause taxonomy that bridges the gap between failure manifestations and actionable repair strategies. 
In addition, we conduct an in-depth investigation into different types of failure root causes and nodes, examining their failure occurrence patterns, underlying causes, and repair difficulty. These analyses together provide a clearer understanding of failure mechanisms in this emerging agent-based system paradigm and yield actionable insights to guide the development of more reliable and robust automated failure attribution and repair methods. Our work also offers practical design guidelines for building more robust agentic workflows, emphasizing modular prompt design, explicit validation, comprehensive check, and progressive design process.



\bibliographystyle{ACM-Reference-Format}
\bibliography{ref}










\end{document}